\pdfoutput=1

\documentclass[11pt]{article}

\usepackage{emnlp2021}

\usepackage{times}
\usepackage{latexsym}

\usepackage[T1]{fontenc}

\usepackage[utf8]{inputenc}

\usepackage{microtype}
\usepackage{hyperref}
\usepackage{url}
\usepackage{graphicx}
\usepackage{amsmath}
\usepackage{amssymb}
\usepackage{algorithm}
\usepackage{algpseudocode}
\usepackage{appendix}
\usepackage{multirow}
\usepackage{setspace}
\usepackage{enumitem}
\usepackage{CJKutf8}
\usepackage{xcolor}
\usepackage{color}
\usepackage{soul}
\usepackage{makecell}
\usepackage{amsmath}

\newcommand{\tabincell}[2]{\begin{tabular}{@{}#1@{}}#2\end{tabular}}  

\usepackage{diagbox}
\usepackage{wrapfig}
\usepackage{amsfonts}

\title{On the Transferability of Adversarial Attacks \\ against Neural Text Classifier}

 \author{
 Liping Yuan$^{1,2}$, Xiaoqing Zheng$^{*1,2}$, Yi Zhou$^{1,2}$\\ 
 \textbf{Cho-Jui Hsieh$^{3}$, Kai-Wei Chang$^{3}$}  \\
  $^1$School of Computer Science, Fudan University, Shanghai, China \\
  $^2$Shanghai Key Laboratory of Intelligent Information Processing \\
  $^3$Department of Computer Science, University of California, Los Angeles, USA \\
  
  \texttt{\{lpyuan19,zhengxq,yizhou17\}@fudan.edu.cn} \\
  \texttt{\{chohsieh,kwchang\}@cs.ucla.edu}
  }

\begin{document}
\maketitle
\begin{abstract}
Deep neural networks are vulnerable to adversarial attacks, where a small perturbation to an input alters the model prediction. In many cases, malicious inputs intentionally crafted for one model can fool another model. In this paper, we present the first study to systematically investigate the transferability of adversarial examples for text classification models and explore how various factors, including network architecture, tokenization scheme, word embedding, and model capacity, affect the transferability of adversarial examples.
Based on these studies, we propose a genetic algorithm to find an ensemble of models that can be used to induce adversarial examples to fool almost all existing models. 
Such adversarial examples reflect the defects of the learning process and the data bias in the training set.
Finally, we derive word replacement rules that can be used for model diagnostics from these adversarial examples.  
\end{abstract}

\section{Introduction}
Recent studies demonstrate that deep neural networks are vulnerable to adversarial examples, intentionally crafted to fool the models.
Although generating adversarial examples for texts has shown to be more challenging than for images due to their discrete nature, many methods have been proposed to generate adversarial text examples and reveal the vulnerability of deep neural networks in natural language processing (NLP) tasks, including reading comprehension \citep{emnlp-17:Jia}, text classification \citep{arxiv-17:Samanta,arxiv-17:Wong,ijcai-18:Liang, emnlp-18:Alzantot,yang2020greedy}, machine translation \citep{iclr-18:ZhaoZhengli,acl-18:Ebrahimi,arxiv-18:Cheng}, dialogue systems \citep{naacl-19:Cheng}, and dependency parsing \citep{acl-20:Zheng}.
These methods perturb text examples by replacing, scrambling, and erasing characters, words or other language units to fool an NLP model. 

\begin{figure}[t]
  \includegraphics[width = 7.7cm]{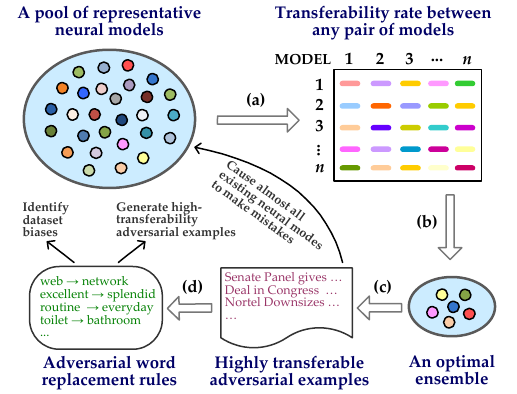}
  \caption{\label{fig:story} Overview of the study.  (a) Given a pool of representative neural models, we compute the adversarial transferability rate between any pair of models; (b) A genetic algorithm is used to find an optimal ensemble with the minimum number of members so that the adversarial examples crafted by attacking the ensemble can strongly transfer to other models; (c) Highly-transferable adversarial examples can be crafted by attacking the ensemble model; (d) We derive adversarial word replacement rules from the adversarial examples constructed by the ensemble such that these rules can be used to identify data biases and to diagnose a model under the black-box setting.}
\end{figure}

\begin{table*} [t]
\small
\setlength{\abovecaptionskip}{0.05cm}
\caption{\label{tb:examples} Three adversarial examples that successfully fool all $63$ models with various configurations (see detailed in Section \ref{sec:configuration}), crafted by using adversarial word replacement rules discovered by our algorithm on AGNEWS.}
\begin{center}
\setlength{\tabcolsep}{1.0mm}
\begin{tabular}{l}

\hline
\hline
\tabincell{l}{Senate Panel Gives NASA Extra Money (AP) AP - NASA would get  \#36;16.4 billion next year under a \\ bill a Senate committee approved Tuesday, reversing a decision by House lawmakers to \textcolor{blue}{\st{cut}} \textcolor{red}{contract} \\ the space agency's budget below this year's levels.} \\

\hline

\tabincell{l}{Deal in Congress to \textcolor{blue}{\st{keep}} \textcolor{red}{preserve} tax cuts, Widening Deficit Republican and Democratic leaders agreed \\ to extend \$5 billion worth of tax cuts sought by \textcolor{blue}{\st{President}} \textcolor{red}{Chairman} Bush without trying to pay for them.} \\

\hline

\tabincell{l}{Nortel Downsizes Again Aug. 23, 2004 (TheDeal.com) Problem-plagued Nortel \textcolor{blue}{\st{Networks}} \textcolor{red}{Web} Corp. \\ announced plans Thursday, Aug. 19, to eliminate an additional 3,500 jobs and fire seven more senior \\ \textcolor{blue}{\st{executives}} \textcolor{red}{administrators} as the company labors to reinvent.} \\

\hline
\hline

\end{tabular}
\end{center}
\vspace{-4mm}
\end{table*}

Most existing studies focus on developing effective algorithms for attacking a specific model. 
The successful attacks demonstrate the instability of model predictions. However, the vulnerability of a model may correlate with different factors, such as network architecture, tokenization scheme, word embedding type, model capacity, and the spurious predictive patterns in the training data.

In this study, we aim to \emph{understand the attack algorithms through the lens of analyzing transferability of adversarial examples.} 
We first systematically investigate which factors of neural models impact the black-box transferability (i.e., how adversarial
examples generated against one model can fool another one~\cite{iclr-14:Szegedy}) of adversarial examples through extensive experiments on two text classification datasets, Sentiment Movie Reviews (MR) \cite{acl-05:Pang}
and AG News corpus (AGNEWS) \cite{nips-15:Zhang}. These factors include network architectures (LSTM, CNN, or Transformer), tokenization schemes (character, sub-word, or word), embedding types (GloVe, word2vec, or fastText), and model capacities (different network depths). 
We vary one factor at a time while fixing  the others to see which factor is the more significant one, and found that the tokenization scheme has the greatest influence on the adversarial transferability, following by network architecture, embedding type, and model capacity.


Based on the analysis, we study whether it is possible to \emph{craft highly-transferable text adversarial examples for many neural models by ensembling a small number of models}. 
Specifically, these highly-transferable adversarial examples provide the following insights. 
First, the adversaries do not need white-box access to victim models. They launch the attacks by their own models trained on similar data, which can transfer across models \citep{cvpr-17:Moosavi-Dezfooli}. 
Second, as stated in \citet{emnlp-19:Wallace}, such adversarial examples are a useful analysis tool and reveal general input-output patterns learned by models, which can be leveraged to study the influence of dataset biases and to identify those biases learned by models.


We also found that the adversarial examples obtained by an ensemble model are more transferable and
propose a genetic algorithm to find an optimal ensemble based on the empirical transferability between different models. 
The adversarial examples generated by attacking the founded ensemble are strongly transferable to other models. For some models, they even exhibit better transferability than those generated by attacking the same model but with different random initialization.

Finally, inspired by \citet{acl-18:Ribeiro}, we generalize the adversarial examples constructed by our ensemble into semantics-preserving adversarial word replacement rules that can induce adversaries on any text input strongly transferring to other neural network-based models (see Table \ref{tb:examples}).
Since those rules are model-agnostic, they provide an analysis of global model behavior and help us to identify dataset biases and to diagnose heuristics learned by the models (See Figure \ref{fig:story} for an illustration of the process).

\section{Adversarial Transferability Among Neural Models}
In the following, we first want to investigate how network architectures, tokenization schemes, embedding types, and model capacities affect the attack transferability.
We conduct an empirical study by varying one factor at a time while fixing the rest to see the differences in their attack transferability.
Technically, we generate the adversarial examples by attacking a source model and pass the generated adversarial examples through other models for comparison. 

\subsection{Experimental Design}
\label{sec:configuration}

We use convolutional neural network (CNN), long short-term memory (LSTM), and bidirectional LSTM as base models with $1$, $2$, and $4$ layers (an additional $6$-layer one for CNN).
Those networks can take three forms as input: word, character, and word + character.
If word-based models are used, their word embeddings are randomly initialized or initialized with GloVe \citep{emnlp-14:Pennington}, word2vec \citep{corr-mikolov:13}, or fastText \citep{arXiv-16:Joulin}.
When taking word + character as input, the models are initialized with the embeddings pre-trained by ELMo \citep{arXiv-18:Peters}. 
We also include BERT \citep{naacl-18:Devlin}, RoBERTa \citep{arXiv-19:Liu}, and ALBERT \citep{arXiv-19:Lan} into the model pool for analysis, and the total number of models under investigation is $63$ (see Appendix A.1 for details), which cover popular neural networks that are used in NLP literature.

All the models are investigated under two recently proposed attack algorithms, PWWS \citep{acl-19:Ren} and GA \citep{emnlp-18:Alzantot}.
The sets of allowed word substitutions are based on the synonyms created in WordNet \citep{acm-95:Miller}, and for any word in a text, the word to replace must have the same part-of-speech (POS) as the original one\footnote{We did not use two recently proposed attack algorithms of BERT-Attack \cite{li2020bert} and BAE \cite{garg2020bae} because they cannot guarantee that any substitute word is always synonymous with the original word.}.
\citet{emnlp-18:Alzantot} also used a language model (LM) to rule out candidate substitute words that do not fit within the context.
However, unlike PWWS, ruling out some candidates by an LM will greatly reduce the number of candidate substitute words ($65\%$ off on average).
For consistency, we report the robust accuracy under GA attack without using an LM.
\citet{acl-20:Zang} suggested that existing textual attack algorithms can roughly be divided into two categories: greedy and population-based algorithms. PWWS and TextFooler \cite{jin2019bert} fall into the first category while GA and PSO \cite{acl-20:Zang} belong to the second one. 
We chose one attack algorithm in each category when investigating the transferability among neural models and use TextFooler to evaluate the generalizability of the proposed method in Section \ref{sec:nli}.

We conducted experiments on two text classification datasets: Sentiment Movie Reviews (MR) \cite{acl-05:Pang}
and AG News corpus (AGNEWS) \cite{nips-15:Zhang}.
All models are trained on the standard training set with the cross-entropy loss.
For each dataset, we attack $1,000$ randomly selected test examples.
For evaluating their transferability on other models, we randomly choose $500$ adversarial examples that successfully cause the source model to make incorrect predictions.
The transferability between each possible pair of  models is shown in Appendix A.2.

\subsection{Significance of Various Factors}
\label{sec:universal-example-generation} 

To find out which factor affects the transferability of adversarial examples the most, we vary one factor at a time while fixing all the others for each model in the pool, and compare the transferability rates between them.
For example, we take a $2$-layer word-based LSTM model randomly initialized, denoted as ``LSTM-Word-Random-$2$'', as a target model.
If we want to know the impact of network architecture, we generate $1,000$ adversarial examples each by attacking BiLSTM-Word-Random-$2$, and CNN-Word-Random-$2$, and use randomly selected $500$ examples each of successful attack to evaluate the robustness of the target model.
If we want to understand the impact of word embedding, the adversarial examples will be crafted by LSTM-Word-GloVe-$2$, LSTM-Word-word2vec-$2$, and LSTM-Word-fastText-$2$ models.

\begin{table}[htbp]
\small
\setlength{\abovecaptionskip}{0.05cm}
\caption{Relative adversarial transferability rate to the base transferability rate on AGNEWS and MR datasets under PWWS and GA attacks.\label{tb:factor} }
\centering
\begin{tabular}{l|cc|cc}

\hline
\hline

\multirow{2}{*}{\bf Transferability} & \multicolumn{2}{c|}{\bf AGNEWS} & \multicolumn{2}{c}{\bf MR} \\ \cline{2-5}
& {\bf PWWS} & {\bf GA} & {\bf PWWS} & {\bf GA} \\

\hline

{\bf Architecture} & $0.197$ & $0.018$ & $0.145$ & $0.021$  \\

{\bf Tokenization} & $0.286$ & $0.030$ & $0.285$ & $0.049$ \\

{\bf Embedding} & $0.085$ & $0.015$ & $0.114$ & $0.015$  \\

{\bf Capacity} & $0.066$ & $0.013$ & $0.045$ & $0.011$ \\

\hline
\hline

\end{tabular}
\end{table}

\begin{table}[htbp]
\small
\setlength{\abovecaptionskip}{0.05cm}
\caption{Adversarial transferability rate among various neural network architectures on AGNEWS dataset under PWWS attack. The architectures listed in the rows are source models, those in the columns are target ones.
\label{tb:architecture}
}
\centering
\begin{tabular}{l|cccc}

\hline
\hline

{\bf Model} & {\bf LSTM} & {\bf BiLSTM} & {\bf CNN} & {\bf BERT}  \\

\hline

{\bf LSTM} & $0.448$ & $0.394$ & $0.353$ & $0.190$ \\

{\bf BiLSTM} & $0.387$ & $0.420$ & $0.337$ & $0.183$  \\

{\bf CNN} & $0.343$ & $0.334$ & $0.442$ & $0.169$ \\

{\bf BERT} & $0.357$ & $0.346$ & $0.348$ & $0.396$ \\

\hline
\hline

\end{tabular}
\end{table}

Since some models may be \emph{inherently} more vulnerable than others, we need to evaluate the \emph{base transferability rate} to remove the effects that are  not caused by the factors we consider.
For each target model, we train two instances with the same setting but different random initialization to obtain its base transferability rate by generating adversarial examples against one model and testing them on another.
This base transferability rate of a model will be subtracted from all the actual adversarial transferability rates obtained when taking the model as the test model\footnote{For example, the base transferability rate of Model A is $90\%$ and that of Model B is $60\%$. If we also know that $70\%$ of adversarial examples produced using Model C can cause Model A to make mistakes, and $50\%$ of those are misclassified by Model B, we may conclude that the transferability rate of C-to-A ($70\%$) is higher than that of C-to-B ($50\%$), which is obviously wrong because the difference between the transferability rate of C-to-B ($50\%$) and B’s base transferability rate ($60\%$) is much smaller than that between the rate of C-to-A ($70\%$) and A’s base rate ($90\%$). We report such (subtracted) adversarial transferability rates in Table \ref{tb:factor} only.}.
We report the average of (subtracted) adversarial transferability rates in Table \ref{tb:factor}, these rates are averaged over the configurations (all possible pairs) described in Section \ref{sec:configuration}. 
Note that the smaller the values are, the more the adversarial transferability rate is close to its base (intra-model) transferability rate.
From these rates, we found the tokenization scheme has the greatest influence on the adversarial transferability, followed by network architecture, embedding type, and model capacity no matter what attack algorithm or dataset is used.



\begin{table*} [ht] 
\small
\caption{\label{tb:input} Adversarial transferability rates with different tokenization schemes and embedding types on AGNEWS dataset under PWWS attack. The models listed in the rows are source models, those in the columns are target ones.
}
\begin{center}
\setlength{\tabcolsep}{1.0mm}
\begin{tabular}{l|ccccccc|c}

\hline
\hline

{\bf Input} & {\bf Random} & {\bf GloVe} & {\bf word2vec} & {\bf fastText} & {\bf Character} & {\bf ELMo} & {\bf BERT} & {\bf Average} \\

\hline

{\bf Random} & $0.457$ & $0.389$ & $0.445$ & $0.434$ & $0.214$ & $0.315$ & $0.166$ & $0.346$  \\ 

{\bf GloVe} & $0.481$ & $0.503$ & $0.489$ & $0.493$ & $0.219$ & $0.336$ & $0.174$ & $0.385$  \\ 

{\bf word2vec} & $0.473$ & $0.413$ & $0.472$ & $0.461$ & $0.216$ & $0.316$ & $0.165$ & $0.360$ \\ 

{\bf fastText} & $0.481$ & $0.442$ & $0.482$ & $0.488$ & $0.222$ & $0.330$ & $0.169$ & $0.373$ \\ 

{\bf Character} & $0.261$ & $0.233$ & $0.256$ & $0.256$ & $0.386$ & $0.300$ & $0.186$ & $0.268$  \\ 

{\bf ELMo} & $0.406$ & $0.379$ & $0.401$ & $0.405$ & $0.256$ & $0.679$ & $0.216$ & $0.392$ \\ 

{\bf BERT} & $0.348$ & $0.329$ & $0.343$ & $0.348$ & $0.328$ & $0.408$ & $0.396$ & $0.357$  \\ 

\hline

{\bf Average} & $0.415$ & $0.384$ & $0.413$ & $0.412$ & $0.263$ & $0.383$ & $0.210$ & $0.354$ \\

\hline
\hline

\end{tabular}
\end{center}
\end{table*}

\subsection{Intra-Factor Transferability}

In the following, we drill down into each specific factor.  
Table \ref{tb:architecture} shows adversarial transferability among different network architectures and configurations.
For example, the architecture of ``BERT'' includes three variants: vanilla BERT, RoBERTa, and ALBERT.
Each cell $(i, j)$ in the table reports the transferability between two classes of models $i$ and $j$.
The value of each cell is computed as follows: for each possible pair of models $(s, t)$ where model $s$ belongs to class $i$ and model $t$ belongs to class $j$, we first calculate the transferability rate between models $s$ and $j$, i.e. the percentage of adversarial examples produced using model $s$ misclassified by model $t$; we then average these transferability rates over all the possible pairs.

As shown in Table \ref{tb:architecture}, the adversarial transferability is not symmetric, i.e.
the transferability of the transfer pair $(i, j)$ might be different from the pair  $(j, i)$.
As expected, intra-model adversarial example transferability rates are consistently higher than inter-model transferability ones.
The adversarial examples generated using BERTs transfer slightly worse than other models whereas BERTs show much more robust to adversarial samples produced using the models from other classes.
It is probably because BERTs were pre-trained with large-scale data and take different tokenization scheme (i.e. sub-words).
We found the models from BERT family tend to distribute their ``attention'' over more words of an input text than others, which makes it harder to change their predictions by perturbing just few words.
In contrast, other models often ``focus'' on certain keywords when making predictions, which makes them more vulnerable to black-box transfer attacks (see Appendix A.3).

In Table \ref{tb:input}, we report the impact of tokenization schemes and embedding types on the adversarial transferability.
Each cell is obtained by the method as the values reported in Table \ref{tb:architecture}.
The pre-trained models show to be more robust against black-box transfer attacks no matter their word embeddings or other parameters or both are pre-trained with large-scale text data.
Character-based models are more robust to transfer attacks than those taking words or sub-words as input, and their adversarial examples also transfer much worse than others.

\subsection{Summary of Findings}
Some findings on the adversarial transferability among models are summarized below: 
\begin{itemize}[leftmargin=*]
\setlength{\itemsep}{0pt}
\setlength{\parsep}{0pt}
\setlength{\parskip}{0pt}
\item No matter what attack algorithm or dataset is used, the tokenization scheme has the greatest impact on the adversarial transferability, followed by the netowrk architecture, embedding type, and model capacity in the order of importance.
\item The adversarial transfer is not symmetric, and the transferability rates of intra-model adversarial examples are consistently higher than those of inter-model ones.
\item Pre-trained neural models show to be more robust against black-box transfer attacks no matter their word embeddings or other parameters or both are pre-trained with large-scale text data.
\item The adversarial examples produced by attacking BERTs transfer slightly worse than others, but BERTs show much more robust to transfer adversarial attacks.
We found that BERTs tend to distribute their ``attention'' over more words than others, which makes it harder to change their predictions by perturbing just few words. Similar observation has been observed in \cite{hsieh2019robustness}. 
\item Character-based models are more robust to transfer attacks than those taking words or sub-words as input, but their adversarial examples also transfer much worse than others.
\item Among the models from BERT family, the models pre-trained with more
data show to be more robust against black-box transfer attacks using the models pre-trained with less data, while the adversarial examples produced by attacking the former transfer slightly better than the latter.
\end{itemize}

We also found that the adversarial examples produced by using an ensemble with a small number of models are much more transferable than those by a single model.
A small ensemble greatly speeds up the adversarial example generation process, which is useful to perform the test-time attacks when evaluating the robustness of local models or launch the online attack in a real-world simulated environment because attacking an ensemble consisting of all possible models is time-consuming and not cost-efficient.
The next two questions are how to select few models from a pool of models to create an ensemble that has good coverage to attack all other models and whether adversarial examples produced by the ensemble can strongly transfer to other unseen models (not listed in the model pool).
We will answer these two questions in Section \ref{sec:highyly_transferable_exmples}.

The above findings can guide us to choose a pool of representative neural models, from which we select a small number of them to form an ensemble.
For example, such a pool of models should include at least one neural network for each type of tokenization scheme,
but it does not need to include too many networks of different depths.
The smaller the number of models in a pool, the less the computational cost will be for estimating the transferability rate between any pair of them.

\section{Highly-Transferable Examples}
\label{sec:highyly_transferable_exmples}

Next, we discuss how to find an optimal ensemble model that can be used to craft adversarial examples that strongly transfer across other models. 
We then distill the ensemble attack into adversarial word replacement rules that can be used to generate adversarial examples with high transferability. These rules can also help us to identify dataset biases and analyze global model behaviors.

\subsection{Ensemble Method}
\label{sec:ensemble-method}

Consider an ensemble model that outputs the prediction score for a class label by averaging over the scores of individual models, we can generate adversarial examples to fool the ensemble model by applying word substitution-based perturbations to input texts.
We take the average of the logits produced by all the member models as the final prediction.
Observing that the transferability is affected by various factors, and many factors need to be carefully considered when forming an ensemble, we propose a population-based genetic algorithm to find an optimal ensemble.

In the proposed algorithm, a candidate solution is a set of models $S = (s_1, s_2, \dots, s_m)$, where $m$ is a pre-defined size of ensemble.
A fitness function evaluates each solution to decide whether it will contribute to the next generation of solutions. 
We define a function $r(s, t, a)$ as the percentage of adversarial samples produced using model $s$ misclassified by model $t$ under attack algorithm $a$. For a solution $S$, the fitness function $f(S)$ that returns a measure of the candidate's fitness which we want to maximize is defined as follows:
\begin{equation}
\small
f(S) \!=\! \sum\limits_{t_j \in T} \left\{\max_{s_i \in S, s_i \neq t_j} \left[\min_{a_k \in A} r(s_i, t_j, a_k)\right]\right\} \, / \, |T|, 
\end{equation}
\noindent where $T$ is a pool of representative models under investigation, $|T|$ is the cardinality of $T$, and $A$ is a set of attack algorithms.
Let $P(n)$ define a population of candidate solutions at the $n$-th generation : $P(n) = \{s^n_1, s^n_2, \dots, s^n_m\}$.
Initial populations $P(0)$ are selected randomly.
After evaluating each candidate by the fintness function, the algorithm takes two candidate solutions based on fitness, merges their sets, and then randomly selects $m$ models from the set to produce new candidates. 
The mutation is another important genetic operator that takes a single candidate and randomly replaces at most one of its models with another one from $T$.
The algorithm continues until the number of generations reaches the maximum value.

In order to evaluate the ensembles found by the population-based algorithm, we ask a senior researcher to select the ensembles as a baseline.
This researcher uses a simple strategy to make selection: first choose the model whose adversarial examples yield the highest transferability, and gradually add complementary models which are different from those already in the ensemble in the aspects of tokenization scheme, architecture, and embedding type.
We list the ensembles selected by the algorithm and the human expert in Appendix A.4.

\begin{figure*}[ht]
  \centering
  \includegraphics[width = 14.5cm]{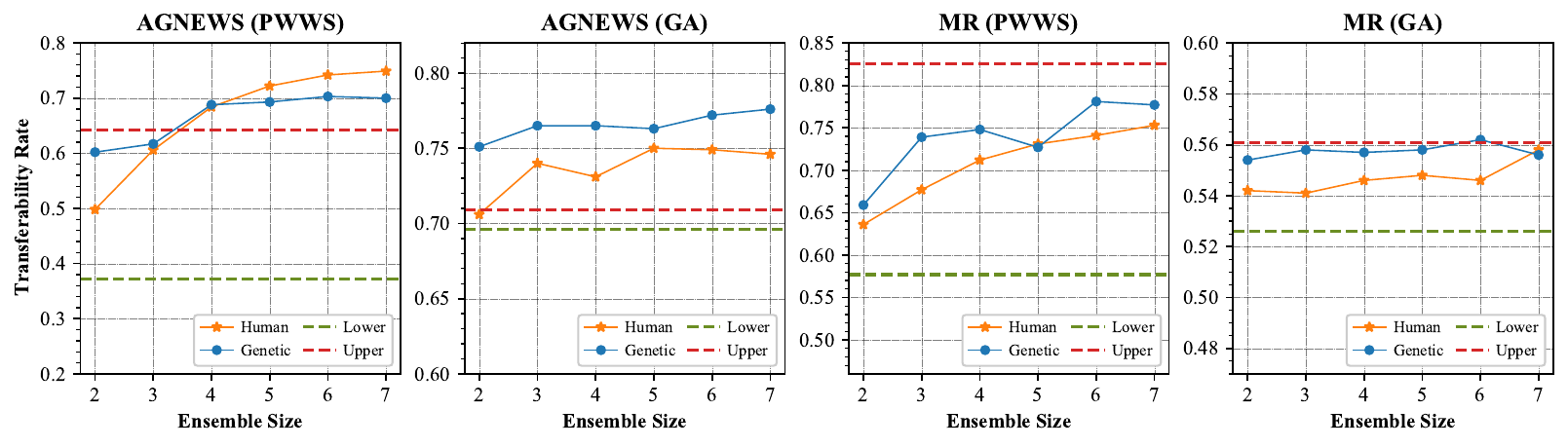}
  \caption{\label{fig:ensemble_comparison} Transferability rates of the adversarial examples generated using different ensembles with various sizes on both AGNEWS and MR datasets under two attack algorithms (PWWS and GA).
  The upper red dotted line represents the average of the \emph{base transferability rates} (defined in Section \ref{sec:universal-example-generation}) that theoretically are highest rates that can be achieved using a single local model, and the lower green line shows the average of transferability rates over all the possible pairs of models.
  The ensemble method clearly outperforms the single model-based transfer method, and 
  in most cases the adversarial examples produced using the ensembles founded by the proposed genetic algorithm transfer better across different models than those selected by a human expert.
  }
\end{figure*}

In Figure \ref{fig:ensemble_comparison}, we show the transferability rates of the adversarial examples produced using the ensembles with various sizes on both AGNEWS and MR datasets under two attack algorithms (PWWS and GA).
The reported transferability rates are averaged over all the remaining models except those used to produce the adversarial examples.
We found that in most cases the adversarial examples produced using the ensemble founded by the  genetic algorithm transfer better across different models than those selected by a human expert, especially when the ensemble size is small.
The ensemble method performs superior to a single model-based transfer method, and in some cases the transferability rates achieved by the ensemble method even go beyond the upper red dotted line (i.e. the highest rates that can be achieved by using a single local model).
When the ensemble size is greater than $6$, the marginal gains in average transferability rate decrease no matter what attack algorithm or dataset is used in our experimental setting.

\begin{figure}[t] \small
 \hangafter 0 \hangindent 0.0em \rule{7.7cm}{0.5pt}

 \begin{spacing}{1.0}

 \hangafter 0 \hangindent 0.5em \textbf{Inputs:}

 \hangafter 0 \hangindent 1em $\mathcal{D}$: a set of training examples.

 
 \hangafter 0 \hangindent 1em $\mathcal{Z}$: a set of class labels.

 \hangafter 0 \hangindent 1em $g$: an ensemble model that outputs a logit for each class 
 
 \hangafter 0 \hangindent 2.2em $z \in \mathcal{Z}$.
 

 \vspace{1.0pt}

 \hangafter 0 \hangindent 0.5em \textbf{Output:} a set of word replacement rules as well as their 
 
 \hangafter 0 \hangindent 4.2em salience.
 
 \vspace{1.0pt}
 

 
 \hangafter 0 \hangindent 0.5em  \textbf{Algorithm:}

 \vspace{1.0pt}

 \hangafter 0 \hangindent 0.5em  $1$: \hspace{2.0pt} \textbf{for} each training instance $(x, y)$ in $\mathcal{D}$

 \vspace{1.0pt}
 
 \hangafter 0 \hangindent 0.5em  $2$: \hspace{10.0pt} \textbf{for} each word $w_i$ in the input text $x$ 
 
 \vspace{1.0pt}
 
 \hangafter 0 \hangindent 0.5em  $3$: \hspace{18.0pt} \textbf{for} each word $\hat{w_i}$ that can be used to replace $w_i$ 
 
 \vspace{1.0pt}
 
 \hangafter 0 \hangindent 0.5em  $4$: \hspace{26.0pt} $\hat{x_i} = $ replace $w_i$ with $\hat{w_i}$ in $x$.
 
 \vspace{1.0pt}
 
 \hangafter 0 \hangindent 0.5em  $6$: \hspace{26.0pt} $c(y, w_i \rightarrow \hat{w_i}) = c(y, w_i \rightarrow \hat{w_i}) + 1$.
 
 \vspace{1.0pt}

 \hangafter 0 \hangindent 0.5em  $5$: \hspace{26.0pt} \textbf{for} each label $z \in \mathcal{Z}$

 \vspace{1.0pt}
 
 \hangafter 0 \hangindent 0.5em  $7$: \hspace{32.0pt} \textbf{if} $z = y$ \textbf{then}
 
 \vspace{1.0pt}
 
 
 \hangafter 0 \hangindent 0.5em  $8$: \hspace{40.0pt} $h(y, w_i \rightarrow \hat{w_i}) = h(y, w_i \rightarrow \hat{w_i}) + $ 
 
 \hangafter 0 \hangindent 13.8em $ g(x_i;z) - g(\hat{x_i};z)$.
 
 \vspace{1.0pt}
 
 \hangafter 0 \hangindent 0.5em  $9$: \hspace{32.0pt} \textbf{else}

 \vspace{1.0pt}
 
 \hangafter 0 \hangindent 0.5em  $10$: \hspace{35.0pt} $h(y, w_i \rightarrow \hat{w_i}) = h(y, w_i \rightarrow \hat{w_i}) + $
 
 \hangafter 0 \hangindent 13.8em $g(\hat{x_i};z) - g(x_i;z)$.

 \vspace{1.0pt}
 
 \hangafter 0 \hangindent 0.5em  $11$: \hspace{1.0pt} \textbf{for} each word replacement rule
 
 \vspace{1.0pt}

 \hangafter 0 \hangindent 0.5em  $12$: \hspace{9.0pt} $h(z, w \rightarrow \hat{w}) = h(z, w \rightarrow \hat{w}) / c(z, w_i \rightarrow \hat{w_i})$.

 \end{spacing}
 \hangafter 0 \hangindent 0.0em \rule{7.7cm}{0.5pt}
 
 \caption{\label{alg:rules}An algorithm to discover highly-transferable adversarial word replacement rules.}
\vspace{-0.5cm}
\end{figure}

\subsection{Mining Word Replacement Rules}
\vspace{-1mm}

We have shown in Section \ref{sec:ensemble-method} that the adversarial examples generated by the ensemble whose members are carefully selected can strongly transfer to other models.
We hypothesize that if we can distill the ensemble attack into some word replacement rules, the adversarial examples crafted by applying the distilled rules to perturb input texts also can transfer well across different models.
In this section, we want to discover such word replacement rules using an ensemble model, and those rules are expected to be used to generate the model-agnostic examples of transferable hostility.
Besides, such rules (if any) also can help us to understand and identify dataset biases ``unknowingly'' exploited by the models for prediction.

A word replacement rule is defined as a pair $(z, w \rightarrow \hat{w})$, where $z$ is a class label, and $w \rightarrow \hat{w}$ means to replace the original word $w$ with $\hat{w}$ when the gold label is $z$. 
Each rule is associated with a salience $h(z, w \rightarrow \hat{w})$ specifying the priority of the rule, and a higher number denotes a higher priority.
We propose an algorithm to discover Highly-transferable Adversarial Word Replacement (HAWR) rules (see Figure \ref{alg:rules}).
The idea behind this algorithm is to estimate the changes in log-likelihood caused by the word replacements.
Once such rules are obtained, they can be used to generate adversarial examples as follows: given an input sentence $x$ and its label $y$, we find a word $w_i$ in $x$ which has the highest value of $h(y, w_i, \hat{w_i})$ and replace $w_i$ with $\hat{w_i}$ in $x$; for all the remaining words in $x$ we repeat the above step until the percentage of words that can be altered reach a given threshold.
Note that such adversarial examples can be generated without access to target models.

\begin{table*} [htbp] \small
\setlength{\abovecaptionskip}{0.05cm}
\caption{\label{tb:univeral} Attack success rates of the adversarial examples generated by applying the word replacement rules found by our algorithm (HAWR) and pointwise mutual information (PMI) on all $63$ and three representative models with AGNEWS and MR datasets, comparing to PWWS and GA attack algorithms. 
``Succ\%'' denotes the attack success rate in terms of the number of sentences, and ``Qry\#'' the average number of queries to the victim model required by the attack algorithms. The maximum percentage of words that are allowed to be perturbed was set to $30\%$.}

\begin{center}
\setlength{\tabcolsep}{0.45mm}
\begin{tabular}{l|cc|cc|cc|cc|cc|cc|cc|cc}

\hline
\hline

\multirow{3}{*}{\tabincell{l}{\bf Success \\  \bf Rate}}
& \multicolumn{8}{c|}{\bf AGNEWS}  
& \multicolumn{8}{c}{\bf MR} \\
\cline{2-17}
 
& \multicolumn{2}{c|}{\bf HAWR}  
& \multicolumn{2}{c|}{\bf PMI} 
& \multicolumn{2}{c|}{\bf PWWS} 
& \multicolumn{2}{c|}{\bf GA}
& \multicolumn{2}{c|}{\bf HAWR}  
& \multicolumn{2}{c|}{\bf PMI} 
& \multicolumn{2}{c|}{\bf PWWS} 
& \multicolumn{2}{c}{\bf GA} \\
\cline{2-17}
 
& {\bf Succ\%}  & {\bf Qry\#} 
& {\bf Succ\%}  & {\bf Qry\#}
& {\bf Succ\%}  & {\bf Qry\#} 
& {\bf Succ\%}  & {\bf Qry\#}
& {\bf Succ\%}  & {\bf Qry\#} 
& {\bf Succ\%}  & {\bf Qry\#}
& {\bf Succ\%}  & {\bf Qry\#} 
& {\bf Succ\%}  & {\bf Qry\#} \\ 

\hline
{\bf ALL} & $69.0$ & $0.0$ & $39.6$ & $0.0$ & $69.1$ & $175.6$ & $82.4$ & $380.6$  & $87.7$ & $0.0$ & $79.0$ & $0.0$ & $92.0$ & $97.6$ & $95.8$ & $130.5$ \\
{\bf LSTM} & $75.9$ & $0.0$ & $45.1$ & $0.0$ & $64.1$ & $175.7$ & $80.7$ & $368.3$  & $95.3$ & $0.0$ & $87.3$ & $0.0$ & $95.2$ & $97.2$ & $97.4$ & $104.3$ \\
{\bf CNN} & $75.1$ & $0.0$ & $39.7$ & $0.0$ & $76.1$ & $174.7$ & $88.7$ & $316.6$  & $89.1$ & $0.0$ & $84.3$ & $0.0$ & $93.5$ & $97.2$ & $96.6$ & $100.2$ \\
{\bf BERT} & $33.0$ & $0.0$ & $19.7$ & $0.0$ & $33.2$ & $178.2$ & $65.5$ & $498.2$  & $58.3$ & $0.0$ & $45.1$ & $0.0$ & $78.5$ & $98.2$ & $92.8$ & $167.4$ \\

\hline
\hline

\end{tabular}
\end{center}
\end{table*}

\begin{table*} [ht] \small
\setlength{\abovecaptionskip}{0.05cm}
\caption{\label{tb:instance} Five adversarial word replacement rules discovered from MR dataset each for the positive and negative categories as well as their changes in the pointwise mutual information (PMI).}
\label{tb:word-replacement-rules}

\begin{center}
\setlength{\tabcolsep}{1.0mm}
\begin{tabular}{c|lcl|c|c}

\hline
\hline

{\bf Class} & \multicolumn{3}{c|}{\bf Word Substitution} & {\bf PMI(Word; Positive)} & {\bf PMI(Word; Negative)} \\

\hline
\multirow{5}{*}{\rotatebox{90}{\bf Positive}}
& flaws & $\rightarrow$ & flaw & $6.408 \rightarrow 2.086$ $\bf (4.392 \downarrow)$ &  $0.000 \rightarrow 6.478$ $\bf (6.408 \uparrow)$ \\

& average & $\rightarrow$ & mediocre & $5.741 \rightarrow 2.086$ $\bf (3.655 \downarrow)$ &  $5.004 \rightarrow 6.408$ $\bf (1.404 \uparrow)$ \\

& glorious & $\rightarrow$ & splendiferous & $6.478 \rightarrow 0.000$ $\bf (6.478 \downarrow)$ &  $0.000 \rightarrow 0.000$ $\bf (0.000 \uparrow)$ \\

& web & $\rightarrow$ & network & $6.478 \rightarrow 0.000$ $\bf (6.478 \downarrow)$ &  $0.000 \rightarrow 6.478$ $\bf (6.478 \uparrow)$ \\

& brilliant & $\rightarrow$ & brainy & $6.168 \rightarrow 0.000$ $\bf (6.168 \downarrow)$ &  $4.109 \rightarrow 6.478$ $\bf (2.369 \uparrow)$ \\

\hline

\multirow{5}{*}{\rotatebox{90}{\bf Negative}}
& toilet & $\rightarrow$ & bathroom & $0.000 \rightarrow 6.478$ $\bf (6.478 \uparrow)$ &  $6.478 \rightarrow 0.000$ $\bf (6.478 \downarrow)$ \\

& excellent & $\rightarrow$ & splendid & $6.013 \rightarrow 6.478$ $\bf (0.466 \uparrow)$ &  $4.620 \rightarrow 0.000$ $\bf (4.620 \downarrow)$ \\

& bizarre & $\rightarrow$ & outlandish & $5.478 \rightarrow 6.478$ $\bf (1.000 \uparrow)$ &  $5.478 \rightarrow 0.000$ $\bf (5.478 \downarrow)$ \\

& excruciating & $\rightarrow$ & harrowing & $0.000 \rightarrow 6.256$ $\bf (6.256 \uparrow)$ & $6.478 \rightarrow 3.671$ $\bf (2.807 \downarrow)$ \\

& routine & $\rightarrow$ & everyday & $2.230 \rightarrow 6.478$ $\bf (4.248 \uparrow)$ &  $6.400 \rightarrow 0.000$ $\bf (6.400 \downarrow)$ \\

\hline
\hline

\end{tabular}
\end{center}
\end{table*}

\begin{table} [htbp] 
\small
\setlength{\abovecaptionskip}{0.05cm}
\caption{\label{tb:snli} Attack success rates on SNLI.
``Succ\%'' and ``Qry\#'' have the same meaning as Tabel \ref{tb:univeral}.
The maximum percentage of words perturbed was set to $30\%$.}

\begin{center}
\setlength{\tabcolsep}{0.7mm}
\begin{tabular}{l|cc|cc}

\hline
\hline

\multirow{2}{*}{\bf Success Rate}
& \multicolumn{2}{c|}{\bf HAWR}  
& \multicolumn{2}{c}{\bf TextFooler} \\

\cline{2-5}

& {\bf Succ\%} & {\bf Qry\#} 
& {\bf Succ\%} & {\bf Qry\#} \\ 

\hline
{\bf ESIM} {\scriptsize \cite{Chen-Qian:2017:ACL}} & $72.4$  & $0.0$  & $72.9$  & $24.8$ \\
{\bf DecomAtt} {\scriptsize \cite{parikh2016decomposable}} & $73.8$ & $0.0$  & $75.8$ & $23.7$ \\
{\bf XLNet} {\scriptsize \cite{jin2019bert}} & $68.4$ & $0.0$  & $67.7$ & $24.4$ \\

\hline
\hline

\end{tabular}
\end{center}
\end{table}

We report the attack success rates of the adversarial examples generated by applying HAWR rules in Table \ref{tb:univeral}.
The attacks based on HAWR rules are comparable to PWWS and GA algorithms that require a large number of queries to the victim model, while the attackers using HAWR do not need to access the victim models. 
We use the ensemble consisting of six models 
founded by our genetic algorithm to discover these HAWR rules in this experiment.
We list five adversarial word replacement rules each for the positive and negative categories discovered from MR dataset in Table \ref{tb:word-replacement-rules}.

To understand these adversarial word replacement rules, we analyze their pointwise mutual information (PMI) between words and class label before and after the replacements.
The PMI of a pair of discrete random variables quantifies the discrepancy between the probability of their coincidence given their joint distribution and their individual distributions.
In this case, it is used to find collocations and associations between words and labels, and the PMI of a word $w$ and a label $z \in \mathcal{Z}$ can be computed as $\text{PMI}(w, z) = p(w, z) / p(w) p(z)$, where $p(\cdot)$ assigns a probability to each possible value.
The results show that the PMI are significantly different for the word and its replacement even though they are synonyms. This demonstrates the data bias in the training data.   

We obtain similar word replacement rules by ranking all the hypothesis words according to their PMI with each label from a training set.
For each label $z \in \mathcal{Z}$ and a possible word replacement $w \rightarrow \hat{w}$, the similar salience of $h(z, w \rightarrow \hat{w})$ can be computed as follows:
\begin{equation}
\small 
\setlength{\abovedisplayskip}{3pt}
\begin{aligned}
    h(z, w \rightarrow \hat{w})=& \left[ \text{PMI}(w,z)-\text{PMI}(\hat{w},z) \right] \\ 
    + \sum\nolimits_{z'\in \mathcal{Z}, z' \neq z} & \left[ \text{PMI}(\hat{w},z')-\text{PMI}(w,z') \right].
\end{aligned}
\setlength{\belowdisplayskip}{3pt}
\end{equation}
\noindent We also report the attack success rates of the adversarial examples generated by the word replacement rules obtained using PMI only in Table \ref{tb:univeral}.
The adversarial examples produced by HAWR rules achieved stronger transferability than those by PMI rules.
We believe that it is because HAWR rules are distilled using the logits predicted by models, and the changes in the logits reflect both the characteristics of neural networks and the contexts in which those word replacements are applied.

\subsection{Case Study: Natural Language Inference}
\label{sec:nli}

To evaluate the generalizability of the proposed method, we redo the entire process (illustrated in Figure \ref{fig:story}) on a new task of natural language inference (NLI) as a case study. 
We conducted the experiments of this task on Stanford Natural Language Inference (SNLI) \cite{emnlp-15:Bowman} dataset.
The HAWR rules generated by our algorithm were tested on three models (ESIM, DecompAtt and XLNet listed in Table \ref{tb:snli}) that are \emph{unseen} during the process of finding these rules, and compared to a recently proposed attack algorithm, called TextFooler \cite{jin2019bert}, which has \emph{not} been used.

We reuse $63$ different neural models for text classification (see Appendix A.1) to create a model pool for this task. 
To perform NLI task, each model in the pool encodes the premise and hypothesis separately and then feeds the concatenation of these encodings to a two-layer feedforward network.
We use the ensemble with six models (see Appendix A.5) identified by our generic algorithm to discover HAWR rules based on the
adversarial transferability rates between any pair of models in the pool.
As shown in Table \ref{tb:snli}, the attacks based on HAWR rules are comparable to TextFooler that requires many queries to the victim models.

\section{Related Work}

\paragraph{Transfer-based Attacks}
Observing that adversarial examples often transfer across different models \citep{iclr-14:Szegedy}, the attackers run standard white-box attacks on local surrogate models to find adversarial examples that are expected to transfer to the target models.
Unfortunately, such a straightforward strategy often suffers from overfitting to specific weaknesses of local models and transfer-based attacks typically have much lower success rates than the attacks directly launched on the target models.
To resolve this problem, many methods have been proposed to improve the transfer success rate of adversarial examples on the target models by perturbing mid-layer activations \citep{eccv-18:Zhou,iccv-19:Huang,iclr-20:Inkawhich,cvpr-20:Wu}, adding regularization terms to the example generation process \citep{cvpr-18:Dong,iccv-19:Huang}, or ensembling multiple local models \citep{nips-18:Wu,iclr-18:Tramer,iclr-17:Liu,emnlp-19:Wallace}.

The proposed ensemble-based methods resemble to 
\citet{iclr-17:Liu}, which hypothesized that if an adversarial example remains adversarial for multiple models, then it is more likely to transfer to other models as well.  
\citet{nips-18:Wu} found that the local non-smoothness of loss surface harms the transferability of adversarial examples, and proposed a variance-reduced attack to enhance the transferability by applying the locally averaged gradient to reduce the local oscillation of loss surface.
The existing studies on the ensemble-based transfer attacks are mainly conducted in the image domain  \cite{nips-18:Wu,iclr-18:Tramer,iclr-17:Liu}, transfer-based attacks for NLP models are relatively much underexplored.


\paragraph{Discovering Adversarial Word Replacement Rules}
\citet{acl-18:Ribeiro} presented semantic-preserving perturbations that cause models to change their predictions by the paraphrases generated using back-translation, and generalized these perturbations into universal replacement rules that induce adversaries on many text instances.
They use the word ``universal'' to mean that their replacement rules can be used to any input text if the rules are matched with the input and these rules were generalized across some specific models.
With a different goal, we aim to find the highly-transferable adversarial replacement rules by which the crafted adversarial examples can fool almost all models.
Besides, the number of their replacement rules is relatively small compared to ours.



\section{Conclusion}
\vspace{-2mm}

We investigated four critical factors of NLP neural models, including network architectures, tokenization schemes, embedding types, and model capacities and how they impact the transferability of text adversarial examples with more than sixty different models.
We also proposed a genetic algorithm to find an optimal ensemble of very few models that can be used to generate adversarial examples that transfer well to all the other models.
Then, we described a algorithm to discover highly-transferable adversarial word replacement rules that can be applied to craft adversarial examples with strong transferability across various neural models without access to any of them.
Finally, since those adversarial examples are model-agnostic, they provide an analysis of global model behavior and help to identify dataset biases.


\section*{Acknowledgements}
This work was partly supported by 
Shanghai Municipal Science and Technology Major Project (No. 2021SHZDZX0103) and National Science Foundation of China (No. 62076068). CJH is supported by NSF IIS-1901527, IIS-2008173 and IIS-2048280. KWC is supported by an Amazon Research Award.
\vspace{-5mm}

\bibliography{anthology}
\bibliographystyle{acl_natbib}

\section*{Appendix}

\appendix

\section*{A.1 All Neural Models under Investigation}
\label{appendix:model_pool}

We systematically investigated many popular architectures of neural models with different configurations. Specifically, we consider various network architectures (LSTM, BiLSTM, CNN, or BERT), tokenization schemes (Word, character, or word + character, denoted by ``W'', ``C'', ``WC'' respectively), word embeddings (randomly-initialized, GloVe, word2vec, or fastText), and model capacities (various numbers of layers). All models under investigation are listed in Table \ref{tb:model_pool}, and we believe that they cover the popular neural networks that have been used for text classification tasks in NLP literature.

\begin{table} [htbp] 
\caption{\label{tb:model_pool} All neural models under investigation.}
\scriptsize
\begin{center}
\setlength{\tabcolsep}{1.0mm}
\begin{tabular}{clclcl}

\hline
\hline

ID & Model & ID & Model \\
\hline

$[1]$  &  LSTM-W-Random-1  & $[2]$  &  LSTM-W-GloVe-1  \\
$[3]$  &  LSTM-W-word2vec-1  & $[4]$  &  LSTM-W-fastText-1  \\
$[5]$  &  LSTM-C-Random-1  & $[6]$  &  LSTM-WC-ELMo-1  \\
$[7]$  &  LSTM-W-Random-2  & $[8]$  &  LSTM-W-GloVe-2  \\
$[9]$  &  LSTM-W-word2vec-2  & $[10]$  &  LSTM-W-fastText-2  \\
$[11]$  &  LSTM-C-Random-2  & $[12]$  &  LSTM-WC-ELMo-2  \\
$[13]$  &  LSTM-W-Random-4  & $[14]$  &  LSTM-W-GloVe-4  \\
$[15]$  &  LSTM-W-word2vec-4  & $[16]$  &  LSTM-W-fastText-4  \\
$[17]$  &  LSTM-C-Random-4  & $[18]$  &  LSTM-WC-ELMo-4  \\
$[19]$  &  BiLSTM-W-Random-1  & $[20]$  &  BiLSTM-W-GloVe-1  \\
$[21]$  &  BiLSTM-W-word2vec-1  &  $[22]$  &  BiLSTM-W-fastText-1  \\
 $[23]$  &  BiLSTM-C-Random-1  &  $[24]$  &  BiLSTM-WC-ELMo-1  \\
 $[25]$  &  BiLSTM-W-Random-2  &  $[26]$  &  BiLSTM-W-GloVe-2  \\
 $[27]$  &  BiLSTM-W-word2vec-2  &  $[28]$  &  BiLSTM-W-fastText-2  \\
 $[29]$  &  BiLSTM-C-Random-2  &  $[30]$  &  BiLSTM-WC-ELMo-2  \\
 $[31]$  &  BiLSTM-W-Random-4  &  $[32]$  &  BiLSTM-W-GloVe-4  \\
 $[33]$  &  BiLSTM-W-word2vec-4  &  $[34]$  &  BiLSTM-W-fastText-4  \\
 $[35]$  &  BiLSTM-C-Random-4  &  $[36]$  &  BiLSTM-WC-ELMo-4  \\
 $[37]$  &  CNN-W-Random-1  &  $[38]$  &  CNN-W-GloVe-1  \\
 $[39]$  &  CNN-W-word2vec-1  &  $[40]$  &  CNN-W-fastText-1  \\
 $[41]$  &  CNN-C-Random-1  &  $[42]$  &  CNN-WC-ELMo-1  \\
 $[43]$  &  CNN-W-Random-2  &  $[44]$  &  CNN-W-GloVe-2  \\
 $[45]$  &  CNN-W-word2vec-2  &  $[46]$  &  CNN-W-fastText-2  \\
 $[47]$  &  CNN-C-Random-2  &  $[48]$  &  CNN-WC-ELMo-2  \\
 $[49]$  &  CNN-W-Random-4  &  $[50]$  &  CNN-W-GloVe-4  \\
 $[51]$  &  CNN-W-word2vec-4  &  $[52]$  &  CNN-W-fastText-4  \\
 $[53]$  &  CNN-C-Random-4  &  $[54]$  &  CNN-WC-ELMo-4  \\
 $[55]$  &  CNN-W-Random-6  &  $[56]$  &  CNN-W-GloVe-6  \\
 $[57]$  &  CNN-W-word2vec-6  &  $[58]$  &  CNN-W-fastText-6  \\
 $[59]$  &  CNN-C-Random-6  &  $[60]$  &  CNN-WC-ELMo-6  \\
 $[61]$  &  BERT  &  $[62]$  &  RoBERTa  \\
 $[63]$  &  ALBERT  &   &    \\

\hline
\hline

\end{tabular}
\end{center}

\end{table}

\section*{A.2 Transferability among Different Neural Models}
\label{append:single_model_trs}

\begin{figure}[htbp] 
 \centering
 \includegraphics[width=7.7cm]{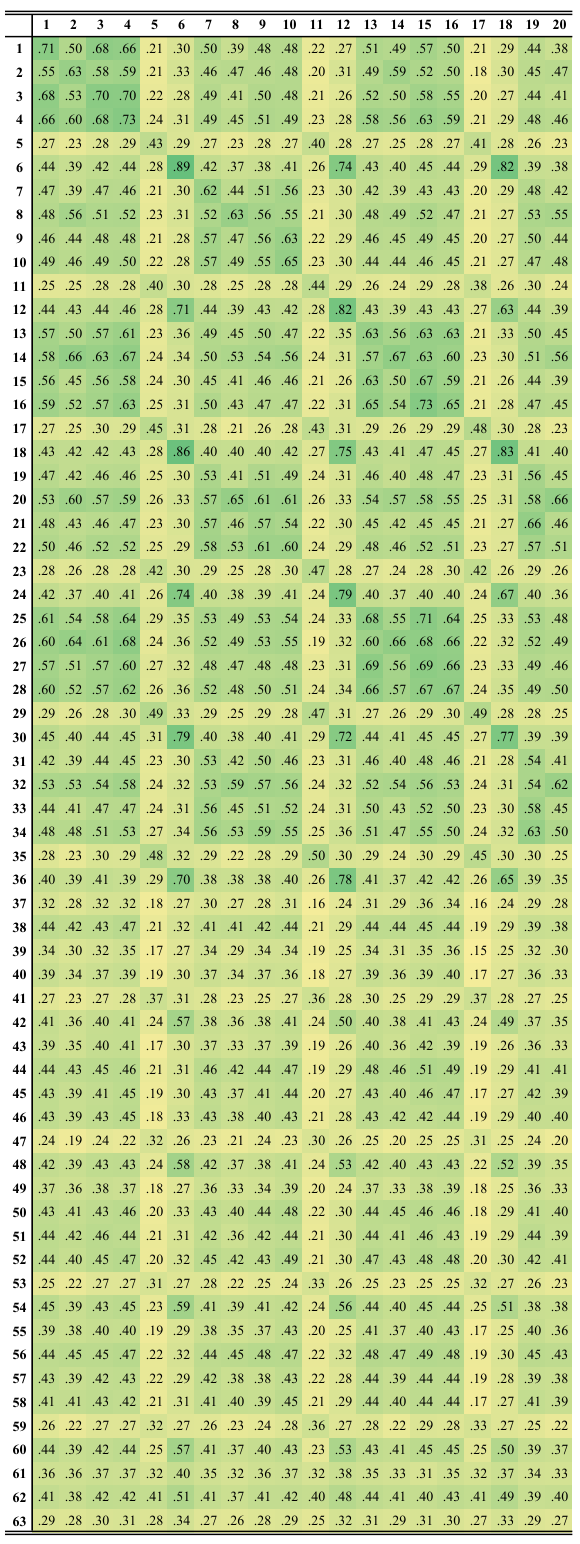}
 \caption{Transferability among neural models (part 1).}
 \label{fig:single_model_tr_agnews_pwws_part1}
\end{figure}

We show in Figure \ref{fig:single_model_tr_agnews_pwws_part1} the transferability rate among all neural models in the model pool.
The column and row headers indicate the IDs of source and target models respectively.
The mapping of IDs and the corresponding models is shown in Figure \ref{tb:model_pool}.
We generate adversarial examples by attacking a source model, and report the transferability rates on a target (or victim) model. 

\addtocounter{figure}{-1}

\begin{figure*}[htbp] 
 \centering
 \includegraphics[width=15cm]{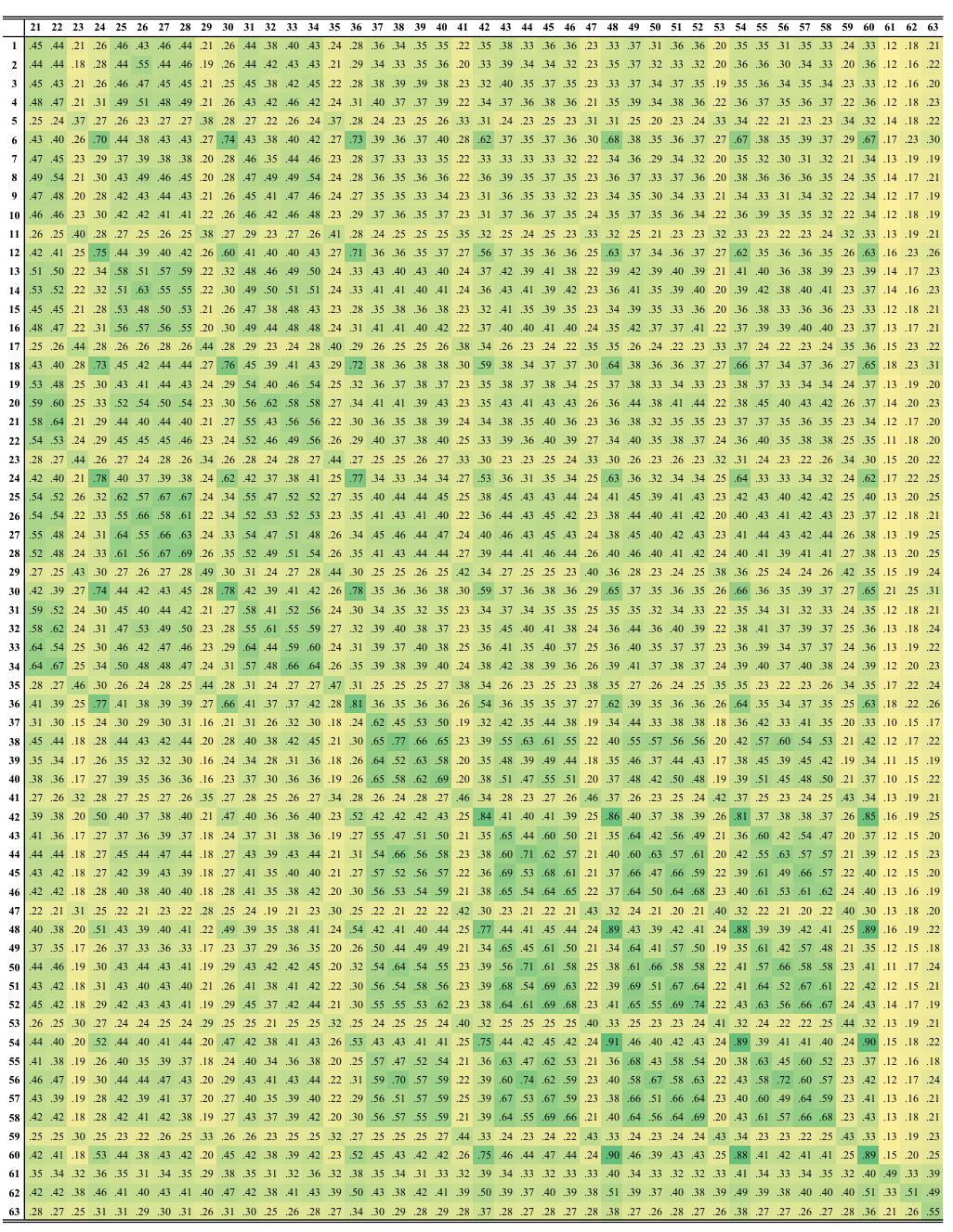}
 \caption{Transferability among neural models (part 2).}
 \label{fig:single_model_tr_agnews_pwws_part2}
\end{figure*}
\addtocounter{figure}{+1}

\begin{figure*}[t]
\centering
\includegraphics[height=7.5cm]{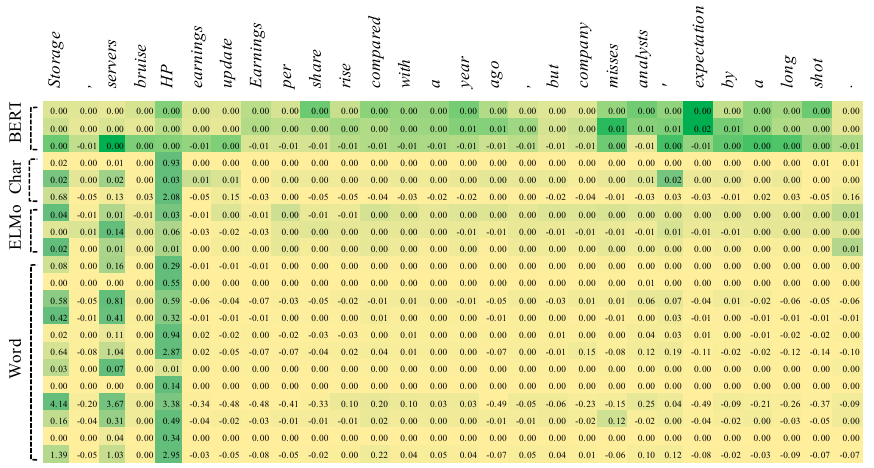}
\caption{Importance of original words.}
\label{fig:attention}
\end{figure*}

\section*{A.3 Heatmap of Word Importance}
\label{appendix:heatmap}

To study how each word in the sentence impacts the prediction of the model, we define word importance as follows:

\begin{itemize}
    \item For an original word, its importance is calculated as the difference between the log likelihood of a gold label before and after the original word is replaced with a special ``unknown'' symbol ($<$unk$>$).
    \item For a substitute word, its importance is estimated as the difference between the log likelihood of a gold label predicted by the model before and after the original word is replaced with the substitute one.
\end{itemize}

\begin{figure}[t]
\centering
\includegraphics[height=8.0cm]{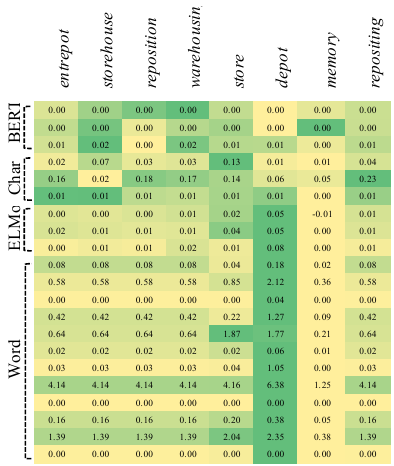}
\caption{Importance of the words used to replace an original word ``Storage'' (the first word in the sentence).}
\label{fig:substitution}
\end{figure}

Figure \ref{fig:attention} and \ref{fig:substitution} show the importance of original and substitute words for different models. We here only consider the models (with one hidden layer) listed in Figure \ref{tb:model_pool} and take the following sentence as an example input:

\textit{Storage, servers bruise HP earnings update Earnings per share rise compared with a year ago, but company misses analysts' expectations by a long shot.}

We observed that different models generally show similar behavior: for the original words, most of the models mainly focus on three words, namely ``Storage'', ``servers'' and ``HP''; for the substitute words, the attentions have been given to the word ``depot'' for most of the models. Thanks to such a similarity, it is possible to generate the adversarial examples using one model, which strongly transfer to the others.

However, the models from BERT family show much more robust to transfer adversarial attacks. 
They tend to distribute their ``attention'' over more words both for original words and substitute words. 
As to character-based models, they also distribute their attention in a way that is clearly different from the other models. 
These differences can explain the lower transferability rates achieved by the adversarial examples generated by using BERTs and character-based models.

\begin{table*} [t] 
\scriptsize
\caption{\label{tb:classification_ensembles} Different ensembles selected by human expert and algorithm on text classification task.}
\begin{center}
\setlength{\tabcolsep}{1.0mm}
\begin{tabular}{c|c|l}
\hline
\hline
\multirow{6}{*}{ \rotatebox{90}{\bf{Human}}}

 & 2 &LSTM-W-Random-1, CNN-C-Random-1 \\
 & 3 &LSTM-W-Random-1, CNN-C-Random-1, LSTM-WC-ELMo-1 \\
 & 4 &LSTM-W-Random-1, CNN-C-Random-1, LSTM-WC-ELMo-1, BERT \\
 & 5 &LSTM-W-Random-1, CNN-C-Random-1, LSTM-WC-ELMo-1, BERT, CNN-W-Random-1 \\
 & 6 &LSTM-W-Random-1, CNN-C-Random-1, LSTM-WC-ELMo-1, BERT, CNN-W-Random-1, LSTM-C-Random-1 \\
 & 7 &LSTM-W-Random-1, CNN-C-Random-1, LSTM-WC-ELMo-1, BERT, CNN-W-Random-1, LSTM-C-Random-1, CNN-WC-ELMo-1 \\
\cline{1-3} 
\multirow{28}{*}{ \rotatebox{90}{\bf{Algorithm}}}
& \multicolumn{2}{c}{\bf{AGNEWS (PWWS)}} \\
\cline{2-3}
 & 2 &LSTM-WC-ELMo-4, CNN-W-GloVe-6 \\
 & 3 &BiLSTM-W-GloVe-2, LSTM-WC-ELMo-4, CNN-W-fastText-4 \\
 & 4 &LSTM-WC-ELMo-1, BiLSTM-W-GloVe-2, CNN-W-fastText-4, RoBERTa \\
 & 5 &LSTM-WC-ELMo-1, BiLSTM-W-GloVe-2, LSTM-W-Random-4, CNN-W-fastText-4, RoBERTa \\
 & 6 &LSTM-WC-ELMo-2, BiLSTM-W-GloVe-2, LSTM-W-Random-4, CNN-W-fastText-4, CNN-WC-ELMo-4, RoBERTa \\
 & 7 &LSTM-WC-ELMo-2, BiLSTM-W-GloVe-2, LSTM-W-Random-4, CNN-W-fastText-4, CNN-WC-ELMo-4, CNN-W-GloVe-6, RoBERTa \\
\cline{2-3}
  & \multicolumn{2}{c}{\bf{AGNEWS (GA)}} \\
\cline{2-3}
 & 2 &LSTM-WC-ELMo-1, CNN-WC-ELMo-1 \\
 & 3 &LSTM-WC-ELMo-1, CNN-C-Random-1, CNN-WC-ELMo-1 \\
 & 4 &BiLSTM-WC-ELMo-4, CNN-C-Random-1, CNN-WC-ELMo-1, CNN-WC-ELMo-6 \\
 & 5 &BiLSTM-WC-ELMo-4, CNN-C-Random-1, CNN-WC-ELMo-1, CNN-WC-ELMo-6, RoBERTa \\
 & 6 &LSTM-WC-ELMo-4, BiLSTM-WC-ELMo-4, CNN-C-Random-1, CNN-WC-ELMo-1, CNN-WC-ELMo-6, RoBERTa \\
 & 7 &LSTM-WC-ELMo-4, BiLSTM-WC-ELMo-4, CNN-C-Random-1, CNN-WC-ELMo-1, CNN-WC-ELMo-2, CNN-WC-ELMo-6, RoBERTa \\
\cline{2-3}
& \multicolumn{2}{c}{\bf{MR (PWWS)}} \\
\cline{2-3}
 & 2 &LSTM-C-Random-4, CNN-WC-ELMo-6 \\
 & 3 &LSTM-C-Random-4, BiLSTM-W-GloVe-4, CNN-WC-ELMo-2 \\
 & 4 &LSTM-C-Random-4, BiLSTM-W-GloVe-4, CNN-W-fastText-2, CNN-WC-ELMo-2 \\
 & 5 &LSTM-W-word2vec-1, LSTM-C-Random-4, CNN-W-fastText-2, CNN-WC-ELMo-2, CNN-W-GloVe-6 \\
 & 6 &LSTM-W-word2vec-1, LSTM-C-Random-4, CNN-W-fastText-2, CNN-WC-ELMo-2, CNN-W-GloVe-6, RoBERTa \\
 & 7 &LSTM-W-word2vec-1, LSTM-C-Random-4, BiLSTM-W-GloVe-4, CNN-WC-ELMo-2, CNN-W-GloVe-6, CNN-W-word2vec-6, RoBERTa \\
\cline{2-3}
 & \multicolumn{2}{c}{\bf{MR (GA)}} \\
\cline{2-3}
 & 2 &BiLSTM-W-GloVe-4, RoBERTa \\
 & 3 &LSTM-C-Random-4, BiLSTM-W-GloVe-4, RoBERTa \\
 & 4 &LSTM-W-Random-1, LSTM-C-Random-4, CNN-W-GloVe-1, RoBERTa \\
 & 5 &LSTM-W-Random-1, LSTM-C-Random-4, BiLSTM-W-GloVe-4, CNN-W-GloVe-1, RoBERTa \\
 & 6 &LSTM-W-Random-1, LSTM-C-Random-4, BiLSTM-W-GloVe-4, BiLSTM-C-Random-4, CNN-W-GloVe-1, RoBERTa \\
 & 7 &LSTM-W-Random-1, LSTM-W-word2vec-1, LSTM-C-Random-4, BiLSTM-W-GloVe-4, BiLSTM-C-Random-4, CNN-W-GloVe-1, RoBERTa \\

\hline
\hline
\end{tabular}
\end{center}
\end{table*}

\begin{table*} [t] 
\scriptsize
\caption{\label{tb:inference_ensembles} Different ensembles selected by genetic algorithm on natural language inference task.}
\begin{center}
\setlength{\tabcolsep}{1.0mm}
\begin{tabular}{c|c|l}
\hline
\hline
\multirow{6}{*}{ \rotatebox{90}{\bf{Ensemble Size}}}

 & 2 & BiLSTM-W-GloVe-2, BiLSTM-WC-ELMo-2 \\
 
 & 3 & RoBERTa, BiLSTM-W-GloVe-2, BiLSTM-WC-ELMo-2 \\
 
 & 4 & RoBERTa, LSTM-W-GloVe-2, CNN-W-fastText-1, BiLSTM-WC-ELMo-2 \\
 
 & 5 & RoBERTa, LSTM-W-GloVe-2, CNN-W-fastText-1, BiLSTM-WC-ELMo-2, BiLSTM-W-GloVe-2 \\
 
 & 6 & RoBERTa, LSTM-W-GloVe-4, CNN-W-fastText-1, BiLSTM-WC-ELMo-2, BiLSTM-W-GloVe-2, LSTM-W-fastText-2 \\
 
 & 7 & RoBERTa, BiLSTM-W-GloVe-2, CNN-W-fastText-1, BiLSTM-WC-ELMo-2, BiLSTM-C-Random-2, LSTM-W-fastText-2, CNN-WC-ELMo-1 \\
\cline{1-3} 

\hline
\hline
\end{tabular}
\end{center}
\end{table*}

\section*{A.4 The Ensembles of Text Classification Task}
\label{appendix:classification_ensembles}

Table \ref{tb:classification_ensembles} shows the ensemble models selected by the proposed  genetic algorithm and human expert of AGNEWS and MR datasets.

\section*{A.5 The Ensembles of Natural Language Inference Task}
\label{appendix:inference_ensembles}

Table \ref{tb:inference_ensembles} shows the ensemble models selected by the proposed  genetic algorithm on SNLI dataset.

\end{document}